# xLSTMTime : Long-term Time Series Forecasting With xLSTM

**Musleh Alharthi * and Ausif Mahmood**

Department of Computer Science and Engineering, University of Bridgeport, Bridgeport, CT 06604, USA; mahmood@bridgeport.edu

**\*** Correspondence: muslehal@my.bridgeport.edu

**Abstract:** In recent years, transformer-based models have gained prominence in multivariate long-term time series forecasting (LTSF), demonstrating significant advancements despite facing challenges such as high computational demands, difficulty in capturing temporal dynamics, and managing long-term dependencies. The emergence of LTSF-Linear, with its straightforward linear architecture, has notably outperformed transformer-based counterparts, prompting a reevaluation of the transformer's utility in time series forecasting. In response, this paper presents an adaptation of a recent architecture termed extended LSTM (xLSTM) for LTSF. xLSTM incorporates exponential gating and a revised memory structure with higher capacity that has good potential for LTSF. Our adopted architecture for LTSF termed as xLSTMTime surpasses current approaches. We compare xLSTMTime's performance against various state-of-the-art models across multiple real-world datasets, demonstrating superior forecasting capabilities. Our findings suggest that refined recurrent architectures can offer competitive alternatives to transformer-based models in LTSF tasks, potentially redefining the landscape of time series forecasting. Code: https://github.com/muslehal/xLSTMTime

## 1. Introduction

Time series forecasting with Artificial Intelligence has been a prominent research area for many years. Historical data on electricity, traffic, finance, and weather are frequently used to train models for various applications. Some of the earlier techniques in time series forecasting relied on statistics and mathematical models like SARIMA [1,2,3], and TBATs [4]. These used moving average and seasonal cycles to capture the patterns for future prediction. With the advent of machine learning, new approaches using Linear Regression [5] were developed. Here, a grouping-based quadratic mean loss function is incorporated to improve linear regression performance in time series prediction. Another approach in machine learning is based on an ensemble of decision trees termed XGBoost [6]. This uses Gradient Boosted Decision Trees (GBDT) where each new tree focuses on correcting the prediction errors of the preceding trees.

Deep learning introduced some newer approaches. Some of the earlier techniques used, Recurrent Neural Networks (RNNs)[7] with varying architectures based on Elman RNN, LSTM (Long Short-Term Memory), and GRU (Gated Recurrent Units). These designs capture sequential dependencies and long-term patterns in the data[8]. The recurrent approaches were followed by use of Convolutional Neural Networks (CNNs) in time series e.g., [9,10,11]. In recent years, transformer-based architectures have become the most popular approach for Natural Language Processing (NLP). Their success in NLP has given rise to the possibility of using them in other domains such as image processing, speech recognition, as well as time series forecasting. Some of the popular transformer-based approached to time series include [12,13,14,15,16,17,18]. Of these, Informer [12] introduces a ProbSparse self-attention mechanism with distillation techniques for efficient key extraction. Autoformer [13] incorporates decomposition and auto-correlation concepts from classic time series analysis. FEDformer [14] leverages a Fourier enhanced structure for linear complexity. One of the recent transformer-based architectures termed as PatchTST [16] breaks down a time series into smaller segments to be used as input tokens for the model. Another recent design iTransformer [18] independently inverts the embedding of each time series variate. The time points of individual series are embedded into variate tokens which are utilized by the attention mechanism to capture multivariate correlations. Further, the feed-forward network is applied for each variate token to learn nonlinear representations. While the above mention designs have shown effective results, transformers face challenges in time series forecasting due to their difficulty in modeling non-linear temporal dynamics, order sensitivity, and high computational complexity for long sequences. Noise susceptibility and handling long-term dependencies further complicate their use in fields involving volatile data such as financial forecasting. Different transformer-based designs such as Autoformer, Informer, and FEDformer aim to mitigate the above issues, but often at the cost of some information loss and interpretability.

As a result, some recent time series research tried to explore approaches other than Transformer based designs. These include LTSF-Linear [19], ELM [20], and Timesnet [21]. LTSF-Linear is extremely simple and uses a single linear layer. It outperforms many Transformer-based models such as Informer, Autoformer, and FEDformer[12,13,14] on the popular time series forecasting benchmarks. TimesNet [20] uses modular TimesBlocks and an inception block to transform 1D time series into 2D, effectively handling variations within and across periods for multi-periodic analysis. ELM further improves the LTSF-Linear by incorporating dual pipelines with batch normalization and reversible instance normalization. With the recent popularity of state-space approaches [22], some research in time series has explored these ideas and have achieved promising results e.g., SpaceTime [23], captures autoregressive processes and includes a "closed-loop" variation for extended forecasting.

The success of LTSF-Linear [19] and ELM [20], with straightforward linear architectures, in outperforming more complex transformer-based models has prompted a reevaluation of approaches to time series forecasting. This unexpected outcome challenges the assumption that increasingly sophisticated architectures necessarily lead to better prediction performance. In light of these findings, we propose enhancements to a recently proposed improved LSTM based architecture termed xLSTM. We adapt and improve xLSTM for time series forecasting and term our architecture as xLSTMTime. This model incorporates exponential gating and a revised memory structure, designed to improve performance and scalability in time series forecasting tasks. We compare our xLSTMTime against various state-of-the-art time series prediction models across multiple real-world datasets, and demonstrate its superior performance highlighting the potential of refined recurrent architectures in this domain.

## 2. Related Work

While LSTM was one of first popular deep learning approaches with applications to NLP, it was over shadowed by the success of transformers. Recently, this architecture was revisited and greatly improved. The revised LSTM is termed as xLSTM - Extended Long Short-Term Memory [24]. It presents enhancements to the traditional LSTM architecture aimed at boosting its performance and scalability for large language models. Key advancements include the introduction of exponential gating for better normalization and stabilization, a revised memory structure featuring scalar and matrix variants, and the integration into residual block backbones. These improvements allow xLSTM to perform competitively with state-of-the-art Transformers [25], and State Space Models [22]. xLSTM has two architecture variations which are termed sLSTM and mLSTM, as explain below.

2.1 sLSTM

The stabilized Long Short-Term Memory (sLSTM) [24] model is an advanced variant of the traditional LSTM architecture that incorporates exponential gating, memory mixing, and stabilization mechanisms. These enhancements improve the model's ability to make effective storage decisions, handle rare token prediction in NLP, capture complex dependencies, and maintain robustness during training and inference. The equations describing sLSTM are as described in [24]. We present these here for the sake of completeness of our work before describing the adaptation of these to the time series forecasting domain.

The architecture of sLSTM is depicted in Figure 1.

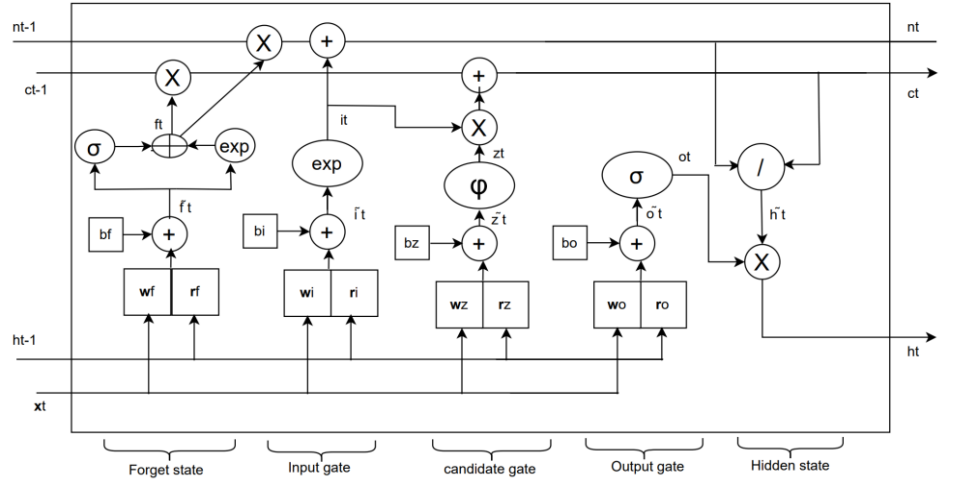

Figure 1: Architecture of sLSTM

For sLSTM, the recurrent relationship between the input and the state is described as:

$$c_t = f_t\, c_{t-1} + i_t z_t \quad (1)$$

where $c_t$ is the cell state at time step $t$. It retains long-term memory of the network, $f_t$ is the forget gate, $i_t$ is the input gate, and $z_t$ controls the amount of input and the previous hidden state $h_{t-1}$ to be added to the cell state, as described below.

$$z_t = \varphi(\tilde{z}_t), \qquad \tilde{Z}_t = \mathcal{W}_z^\intercal x_t + r_z h_{t-1} + b_z \quad (2)$$

In above equations, $x_t$ is input vector, $\varphi$ is an activation function, $\mathcal{W}_z^\intercal$ is the weight matrix, $r_z$ is the recurrent weight matrix, and . $b_z$ represents bias.

The model also uses a normalization state as:

$$n_t = f_t\, n_{t-1} + i_t \quad (3)$$

where $n_t$ is the normalized state at time step $t$. It helps in normalizing the cell state updates. Hidden state $h_t$ is used for recurrent connections as:

$$h_t = o_t\, \tilde{h}_t, \qquad \tilde{h}_t = c_t / n_t \quad (4)$$

where $o_t$ is the output gate. The input gate $i_t$ controls the extent to which new information is added to the cell state as:

$$i_t = \exp(\tilde{\iota}_t), \qquad \tilde{\iota}_t = \mathcal{W}_z^\intercal x_t + r_i h_{t-1} + b_i \quad (5)$$

Similarly the forget gate $f_t$ controls the extent to which the previous cell state $c_{t-1}$ is retained.

$$f_t = \sigma(\tilde{f}_t) \text{ OR } exp(\tilde{f}_t) \qquad \tilde{f}_t = \mathcal{W}_f^\intercal x_t + r_f h_{t-1} + b_f \qquad (6)$$

The output gate $o_t$ controls the flow of information from the cell state to the hidden state as:

$$o_t = \sigma(\tilde{o}_t) \quad , \qquad \tilde{o}_t = \mathcal{W}_o^\intercal x_t + r_o h_{t-1} + b_o \qquad (7)$$

where $\mathcal{W}_o^\intercal$ is the weight matrix that is applied to the current input $x_t$, $r_o$ is the recurrent weight matrix for the output gate that is applied to the previous hidden state $h_{t-1}$ and $b_o$ is the bias term for the output gate.

To provide numerical stability for exponential gates, the forget and input gates are combined into another state $m_t$ as:

$$m_t = \max(\log(f_t) + m_{t-1}, \log(i_t)) \qquad (8)$$

$$i'_t = \exp(\log(i_t) - m_t) = \exp(\tilde{\imath}_t - m_t) \qquad (9)$$

Where $i'_t$ is stabilized input gate which is a rescaled version of the original input gate. Similarly, forget gate is stabilized via $f'_t$ which is a rescaled version of the original forget gate as:

$$f'_t = \exp(\log(f_t) + m_{t-1} - m_t) \qquad (10)$$

To summarize, compared to the original LSTM, the sLSTM adds exponential gating as indicated by equations 5 and 6. Further, use of normalization via equation 3, and finally the stabilization achieved via equations, 8, 9 and 10. These provide considerable improvements to the canonical LSTM.

2.2 mLSTM

The Matrix Long Short-Term Memory (mLSTM) model [24] introduces a matrix memory cell along with a covariance update mechanism for key-value pair storage which significantly increases the model's memory capacity. The gating mechanisms work in tandem with the covariance update rule to manage memory updates efficiently. By removing hidden-to-hidden connections, mLSTM operations can be executed in parallel, which speeds up both training and inference processes. These improvements make mLSTM highly efficient for storing and retrieving information, making it ideal for sequence modeling tasks that require substantial memory capacities, such as language modeling, speech recognition, and time series forecasting. mLSTM represents a notable advancement in recurrent neural networks, addressing the challenges of complex sequence modeling effectively. Figure 2 shows the architecture of mLSTM.

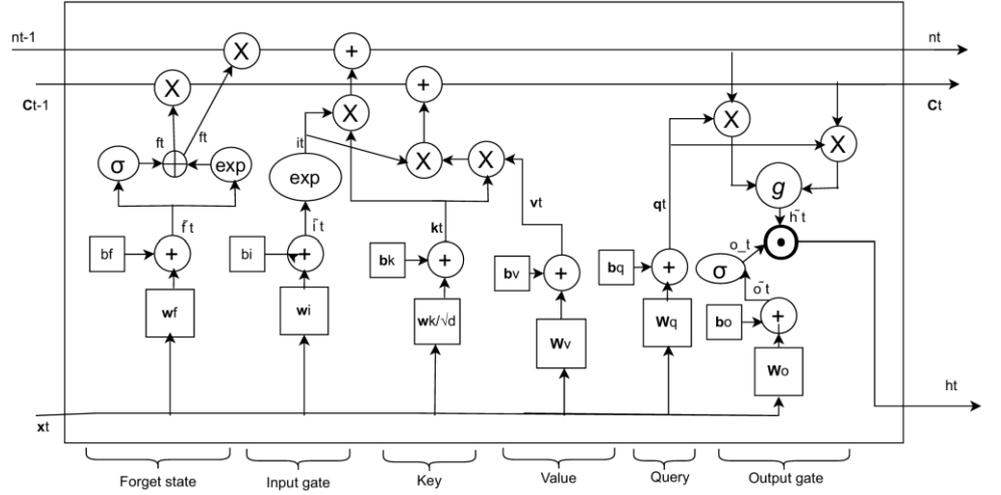

Figure 2: Architecture of mLSTM

Equations 11-19 describe the operations of mLSTM [24].

$$C_t = f_t C_{t-1} + i_t v_t k_t \quad (11)$$

$C_t$ is the matrix memory that stores information in a more complex structure than the scalar cell state in a traditional LSTM. Normalization is carried out similar to sLSTM as:

$$n_t = f_t n_{t-1} + i_t k_t \quad (12)$$

$$h_t = o_t \odot \tilde{h}_t \, , \, \tilde{h}_t = g(C_t, q_t, n_t) = C_t q_t / \max\{n_t^\top q_t, 1\} \quad (13)$$

Similar to the transformer architecture, query $q_t$, key $k_t$, and value $v_t$ are created as:

$$q_t = W_q x_t + b_q \quad (14)$$

$$k_t = \frac{1}{\sqrt{d}} W_k x_t + b_k \quad (15)$$

$$v_t = W_v x_t + b_v \quad (16)$$

$$i_t = \exp(\tilde{i}_t), \qquad \tilde{i}_t = w_i x_t + b_i \quad (17)$$

where $i_t$ is the input gate that controls the incorporation of new information into the memory. The forget gate is slightly different as compared to sLSTM as shown below. It determines how much of the previous memory $C_{t-1}$ is to be retained.

$$f_t = \sigma(\tilde{f}_t) \; OR \; exp(\tilde{f}_t) \, , \qquad \tilde{f}_t = w_f x_t + b_f \quad (18)$$

The output gate is also slightly different in mLSTM as shown below.

$$o_t = \sigma(\tilde{o}_t) \, , \qquad \tilde{o}_t = w_o x_t + b_o \quad (19)$$

The output gate controls how much of the retrieved memory is passed to the hidden state.

In the next section, we describe how we adapt the sLSTM and mLSTM to the time series domain.

## 3. Proposed Method

Our proposed xLSTMTime based model combines several key components to effectively handle time series forecasting tasks. Figure 3 provides an overview of the model architecture.

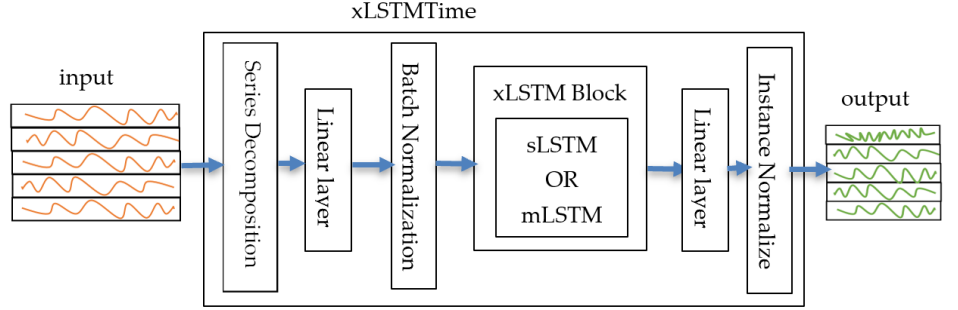

**Figure 3.** xLSTMTime - Data Processing pipeline for xLSTM-based Model for Time Series Forecasting

The input to the model is a time series comprising of multiple sequences. The Series Decomposition block splits the input time series data into two components for each series to capture trend and seasonal information. We implement the approach as proposed in [13] and described as follows. For the input sequence with context length of $L$ and $m$ number of features, i.e., $x \in \mathbb{R}^{L \times m}$, we apply learnable moving averages on each feature via 1-D convolutions. Then the trend and seasonal components are extracted as:

$$x_{trend} = AveragePool(Padding(x))$$
$$x_{seasonal} = x - x_{trend}$$
(20)

After decomposition, the data passes through a linear transformation layer to transform it to the dimensionality needed for the xLSTM modules. We further do a batch normalization [26] to provide stability in learning before feeding the data to the xLSTM modules. Batch Normalization is a transformative technique in deep learning that stabilizes the distribution of network inputs by normalizing the activations of each layer. It allows for higher learning rates, accelerates training, and reduces the need for strict initialization and some forms of regularization like Dropout. By addressing internal covariate shift, Batch Normalization improves network stability and performance across various tasks. It introduces minimal overhead with two additional trainable parameters per layer, enabling deeper networks to train faster and more effectively. [26]

The xLSTM block contains both the sLSTM and the mLSTM components. The sLSTM component uses scalar memory and exponential gating to manage long-term dependencies and controlling the appropriate memory for the historical information. The mLSTM component uses matrix memory and a covariance update rule to enhance storage capacity and relevant information retrieval capabilities. Depending upon the attributes of the dataset, we choose either the sLSTM or the mLSTM component. For smaller datasets such as ETTm1, ETTm2, ETTh1, ETTh2, ILI and weather, we use sLSTM, whereas for larger datasets such as Electricity, Traffic, and PeMS , the mLSTM is chosen due to its higher memory capacity in better learning for time series patterns. The output from the xLSTM block goes through another linear layer. This layer further transforms the data, preparing it for the final output via instance normalization. The Instance Normalization operates on each channel of time series independently. It normalizes the data within each channel of each component series to have a mean of 0 and a variance of 1. The formula for instance normalization for a given feature map is as follows:

$$IN(x) = \frac{x - \mu(x)}{\sigma(x)} \tag{21}$$

where x represents the input feature map, μ(x) is the mean of the feature map and σ(x) is the standard deviation of the feature map [27].

## 4. Results

We test our proposed xLSTM-based architecture on 12 widely used datasets from real-world applications. These datasets include the Electricity Transformer Temperature (ETT) series, which are divided into ETTh1 and ETTh2 (hourly intervals), and ETTm1 and ETTm2 (5-minute intervals). Additionally, we analyze datasets related to Traffic (hourly), Electricity (hourly), Weather (10-minute intervals), and Influenza-Like Illness (ILI) (weekly). Another dataset PeMS (PEMS03, PEMS04, PEMS07 and PEMS08) traffic is sourced from the California Transportation Agencies (CalTrans) Performance Measurement System (PeMS).

**Table 1.** Characteristics of the different datasets used.

| Datasets | Features | Timesteps | Granularity |
|---|---|---|---|
| Weather | 21 | 52,696 | 10 min |
| Traffic | 862 | 17,544 | 1 h |
| ILI | 7 | 966 | 1 week |
| ETTh1/ ETTh2 | 7 | 17,420 | 1 h |
| ETTm1/ETTm2 | 7 | 69,680 | 5 min |
| PEMS03 | 358 | 26,209 | 5 min |
| PEMS04 | 307 | 16,992 | 5 min |
| PEMS07 | 883 | 28,224 | 5 min |
| PEMS08 | 170 | 17,856 | 5 min |
| Electricity | 321 | 26,304 | 1 h |

Each model follows a consistent experimental setup, with prediction lengths T of {96, 192, 336, 720} for all datasets except the ILI dataset. For the ILI dataset, we use prediction lengths of {24, 36, 48, 60}. The look-back window *L* is 512 for all datasets except ILI dataset, for which we use *L* of 96 [16]. We use Mean Absolute Error (MAE) during the training. For evaluation, the metrics used are MSE (Mean Squared Error) and MAE (Mean Absolute Error). Table 2 presents the results for the different benchmarks, comparing our results to the recent works in the time series field.

**Table 2.** Comparison of our xLSTMTime model with other models on the time series datasets.

| Models | xLSTMTime | PatchTST | DLinear | FEDformer | Autoformer | Informer | Pyraformer |
|---|---|---|---|---|---|---|---|

| | Metric | MSE | MAE | MSE | MAE | MSE | MAE | MSE | MAE | MSE | MAE | MSE | MAE | MSE | MAE |
|---|---|---|---|---|---|---|---|---|---|---|---|---|---|---|---|
| Weather | 96 | 0.144 | 0.187 | 0.149 | 0.198 | 0.176 | 0.237 | 0.238 | 0.314 | 0.249 | 0.329 | 0.354 | 0.405 | 0.896 | 0.556 |
| | 192 | 0.192 | 0.236 | 0.194 | 0.241 | 0.220 | 0.282 | 0.275 | 0.329 | 0.325 | 0.370 | 0.419 | 0.434 | 0.622 | 0.624 |
| | 336 | 0.237 | 0.272 | 0.245 | 0.282 | 0.265 | 0.319 | 0.339 | 0.377 | 0.351 | 0.391 | 0.583 | 0.543 | 0.739 | 0.753 |
| | 720 | 0.313 | 0.326 | 0.314 | 0.334 | 0.323 | 0.362 | 0.389 | 0.409 | 0.415 | 0.426 | 0.916 | 0.705 | 1.004 | 0.934 |
| Traffic | 96 | 0.358 | 0.242 | 0.360 | 0.249 | 0.410 | 0.282 | 0.576 | 0.359 | 0.597 | 0.371 | 0.733 | 0.410 | 2.085 | 0.468 |
| | 192 | 0.378 | 0.253 | 0.379 | 0.256 | 0.423 | 0.287 | 0.610 | 0.380 | 0.607 | 0.382 | 0.777 | 0.435 | 0.867 | 0.467 |
| | 336 | 0,392 | 0,261 | 0.392 | 0.264 | 0.436 | 0.296 | 0.608 | 0.375 | 0.623 | 0.387 | 0.776 | 0.434 | 0.869 | 0.469 |
| | 720 | 0,434 | 0,287 | 0.432 | 0.286 | 0.466 | 0.315 | 0.621 | 0.375 | 0.639 | 0.395 | 0.827 | 0.466 | 0.881 | 0.473 |
| Electricity | 96 | 0.128 | 0.221 | 0.129 | 0.222 | 0.14 | 0.237 | 0.186 | 0.302 | 0.196 | 0.313 | 0.304 | 0.393 | 0.386 | 0.449 |
| | 192 | 0.150 | 0.243 | 0.147 | 0.240 | 0.153 | 0.249 | 0.197 | 0.311 | 0.211 | 0.324 | 0.327 | 0.417 | 0.386 | 0.443 |
| | 336 | 0.166 | 0.259 | 0.163 | 0.259 | 0.169 | 0.267 | 0.213 | 0.328 | 0.214 | 0.327 | 0.333 | 0.422 | 0.378 | 0.443 |
| | 720 | 0.185 | 0.276 | 0.197 | 0.290 | 0.203 | 0.301 | 0.233 | 0.344 | 0.236 | 0.342 | 0.351 | 0.427 | 0.376 | 0.445 |
| Illness | 24 | 1.514 | 0.694 | 1.319 | 0.754 | 2.215 | 1.081 | 2.624 | 1.095 | 2.906 | 1.182 | 4.657 | 1.449 | 1.420 | 2.012 |
| | 36 | 1.519 | 0.722 | 1.579 | 0.870 | 1.963 | 0.963 | 2.516 | 1.021 | 2.585 | 1.038 | 4.650 | 1.463 | 7.394 | 2.031 |
| | 48 | 1.500 | 0.725 | 1.553 | 0.815 | 2.130 | 1.024 | 2.505 | 1.041 | 3.024 | 1.145 | 5.004 | 1.542 | 7.551 | 2.057 |
| | 60 | 1.418 | 0.715 | 1.470 | 0.788 | 2.368 | 1.096 | 2.742 | 1.122 | 2.761 | 1.114 | 5.071 | 1.543 | 7.662 | 2.100 |
| ETTh1 | 96 | 0.368 | 0.395 | 0.370 | 0.400 | 0.375 | 0.399 | 0.376 | 0.415 | 0.435 | 0.446 | 0.941 | 0.769 | 0.664 | 0.612 |
| | 192 | 0.401 | 0.416 | 0.413 | 0.429 | 0.405 | 0.416 | 0.423 | 0.446 | 0.456 | 0.457 | 1.007 | 0.786 | 0.790 | 0.681 |
| | 336 | 0.422 | 0.437 | 0.422 | 0.440 | 0.439 | 0.443 | 0.444 | 0.462 | 0.486 | 0.487 | 1.038 | 0.784 | 0.891 | 0.738 |
| | 720 | 0.441 | 0.465 | 0.447 | 0.468 | 0.472 | 0.490 | 0.469 | 0.492 | 0.515 | 0.517 | 1.144 | 0.857 | 0.963 | 0.782 |
| ETTh2 | 96 | 0.273 | 0.333 | 0.274 | 0.337 | 0.289 | 0.353 | 0.332 | 0.374 | 0.332 | 0.368 | 1.549 | 0.952 | 0.645 | 0.597 |
| | 192 | 0.340 | 0.378 | 0.341 | 0.382 | 0.383 | 0.418 | 0.407 | 0.446 | 0.426 | 0.434 | 3.792 | 1.542 | 0.788 | 0.683 |
| | 336 | 0.373 | 0.403 | 0.329 | 0.384 | 0.448 | 0.465 | 0.400 | 0.447 | 0.477 | 0.479 | 4.215 | 1.642 | 0.907 | 0.747 |
| | 720 | 0.398 | 0.430 | 0.379 | 0.422 | 0.605 | 0.551 | 0.412 | 0.469 | 0.453 | 0.490 | 3.656 | 1.619 | 0.963 | 0.783 |
| ETTm1 | 96 | 0.286 | 0.335 | 0.293 | 0.346 | 0.299 | 0.343 | 0.326 | 0.390 | 0.510 | 0.492 | 0.626 | 0.560 | 0.543 | 0.510 |
| | 192 | 0.329 | 0.361 | 0.333 | 0.370 | 0.335 | 0.365 | 0.365 | 0.415 | 0.514 | 0.495 | 0.725 | 0.619 | 0.557 | 0.537 |
| | 336 | 0.358 | 0.379 | 0.369 | 0.392 | 0.369 | 0.386 | 0.392 | 0.425 | 0.510 | 0.492 | 1.005 | 0.741 | 0.754 | 0.655 |
| | 720 | 0.416 | 0.411 | 0.416 | 0.420 | 0.425 | 0.421 | 0.446 | 0.458 | 0.527 | 0.493 | 1.133 | 0.845 | 0.908 | 0.724 |
| ETTm2 | 96 | 0.164 | 0.250 | 0.166 | 0.256 | 0.167 | 0.260 | 0.180 | 0.271 | 0.205 | 0.293 | 0.355 | 0.462 | 0.435 | 0.507 |
| | 192 | 0.218 | 0.288 | 0.223 | 0.296 | 0.224 | 0.303 | 0.252 | 0.318 | 0.278 | 0.336 | 0.595 | 0.586 | 0.730 | 0.673 |
| | 336 | 0.271 | 0.322 | 0.274 | 0.329 | 0.281 | 0.342 | 0.324 | 0.364 | 0.343 | 0.379 | 1.270 | 0.871 | 1.201 | 0.845 |
| | 720 | 0.361 | 0.380 | 0.362 | 0.385 | 0.397 | 0.421 | 0.410 | 0.420 | 0.414 | 0.419 | 3.001 | 1.267 | 3.625 | 1.451 |

Table 2: Multivariate long-term forecasting outcomes with prediction intervals T = {24, 36, 48, 60} for the ILI dataset and T = {96, 192, 336, 720} for other datasets. The best results are highlighted in red, and the next best results are in blue. The lower number is better.

As can be seen from Table 2, for a vast majority of the benchmarks, we outperform existing approaches. Only in case of Electricity and ETTh2, in a few of the prediction lengths, our results are second best.

Figures 4 and 5 show the graphs for actual versus predicted time series values for a few of the datasets. As can be seen, our model learns the periodicity and the variations in the data very nicely for the most part.

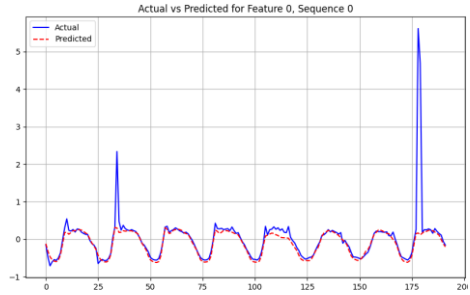
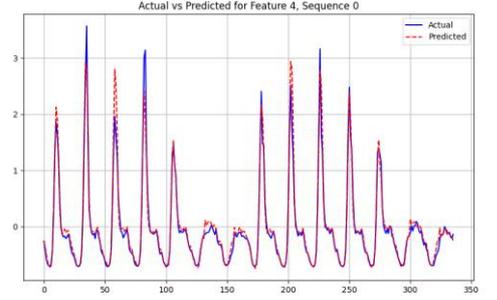

(**a**) T = 192  (**b**) T=336

**Figure 4.** Predicted vs. actual forecasting using our model with L= 512 and T = {192, 336} for traffic dataset.

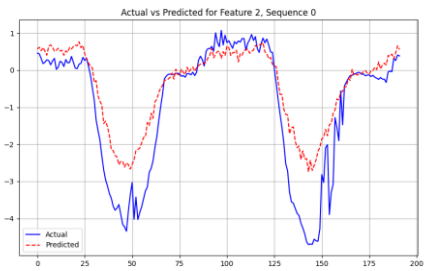
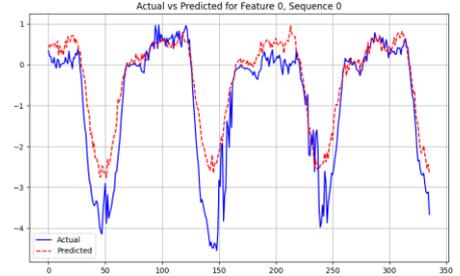

(**a**) T = 192  (**b**) T=336

**Figure 5.** Predicted vs. actual forecasting using our model with L= 512 and T = {192, 336} for ettm1 dataset.

Table 3 presents the comparison results for the PeMS datasets. Here our model produces either the best or the second best results when compared with the recent state of the art models. Figure 6 shows the actual versus predicted graphs for some of the PeMS datasets.

**Table 3.** Comparison of our xLSTMTime model with other models on PEMS datasets. Multivariate forecasting outcomes with prediction intervals T = {12, 24, 48, 96} for all datasets, look-back window L= 96. The best results are highlighted in red, and the next best results are in blue. The lower number is better.

| | Models | Our xlstmTime | | iTransformer | | RLinear | | PatchTST | | Crossformer | | DLinear | | SCINet | |
|---|---|---|---|---|---|---|---|---|---|---|---|---|---|---|---|
| | Metric | MSE | MAE | MSE | MAE | MSE | MAE | MSE | MAE | MSE | MAE | MSE | MAE | MSE | MAE |
| PEMS03 | 12 | 0.065 | 0.166 | 0.071 | 0.174 | 0.126 | 0.236 | 0.099 | 0.216 | 0.090 | 0.203 | 0.122 | 0.243 | **0.066** | **0.172** |
| | 24 | 0.087 | 0.194 | 0.093 | 0.201 | 0.246 | 0.334 | 0.142 | 0.259 | 0.121 | 0.240 | 0.201 | 0.317 | **0.085** | **0.198** |
| | 48 | **0.125** | **0.232** | 0.125 | 0.236 | 0.551 | 0.529 | 0.211 | 0.319 | 0.202 | 0.317 | 0.333 | 0.425 | 0.127 | 0.238 |
| | 96 | 0.192 | 0.291 | 0.071 | 0.174 | 0.126 | 0.236 | 0.099 | 0.216 | 0.090 | 0.203 | 0.457 | 0.515 | 0.178 | 0.287 |
| PEMS04 | 12 | 0.074 | 0.175 | 0.078 | 0.183 | 0.138 | 0.252 | 0.105 | 0.224 | 0.098 | 0.218 | 0.148 | 0.272 | **0.073** | **0.177** |
| | 24 | 0.090 | 0.195 | 0.095 | 0.205 | 0.258 | 0.348 | 0.153 | 0.275 | 0.131 | 0.256 | 0.224 | 0.340 | **0.084** | **0.193** |
| | 48 | 0.123 | 0.230 | 0.120 | 0.233 | 0.572 | 0.544 | 0.229 | 0.339 | 0.205 | 0.326 | 0.355 | 0.437 | **0.099** | **0.211** |
| | 96 | 0.174 | 0.280 | 0.150 | 0.262 | 1.137 | 0.820 | 0.291 | 0.389 | 0.402 | 0.457 | 0.452 | 0.504 | **0.114** | **0.227** |
| PEMS07 | 12 | **0.059** | **0.151** | 0.067 | 0.165 | 0.118 | 0.235 | 0.095 | 0.207 | 0.094 | 0.200 | 0.115 | 0.242 | 0.068 | 0.171 |
|  | 24 | **0.077** | **0.170** | 0.088 | 0.190 | 0.242 | 0.341 | 0.150 | 0.262 | 0.139 | 0.247 | 0.210 | 0.329 | 0.119 | 0.225 |
|  | 48 | **0.105** | **0.204** | 0.110 | 0.215 | 0.562 | 0.541 | 0.253 | 0.340 | 0.311 | 0.369 | 0.398 | 0.458 | 0.149 | 0.237 |
|  | 96 | 0.148 | 0.247 | **0.139** | **0.245** | 1.096 | 0.795 | 0.346 | 0.404 | 0.396 | 0.442 | 0.594 | 0.553 | 0.141 | 0.234 |
| PEMS08 | 12 | **0.072** | **0.169** | 0.079 | 0.182 | 0.133 | 0.247 | 0.168 | 0.232 | 0.165 | 0.214 | 0.154 | 0.276 | 0.087 | 0.184 |
|  | 24 | **0.101** | **0.199** | 0.115 | 0.219 | 0.249 | 0.343 | 0.224 | 0.281 | 0.215 | 0.260 | 0.248 | 0.353 | 0.122 | 0.221 |
|  | 48 | **0.149** | **0.238** | 0.186 | 0.235 | 0.569 | 0.544 | 0.321 | 0.354 | 0.315 | 0.355 | 0.440 | 0.470 | 0.189 | 0.270 |
|  | 96 | 0.224 | 0.289 | **0.221** | **0.267** | 1.166 | 0.814 | 0.408 | 0.417 | 0.377 | 0.397 | 0.674 | 0.565 | 0.236 | 0.300 |

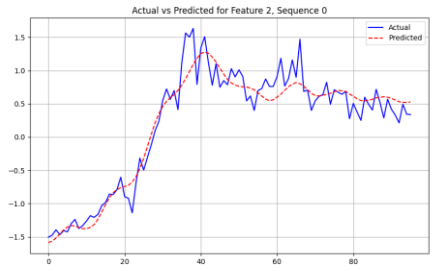
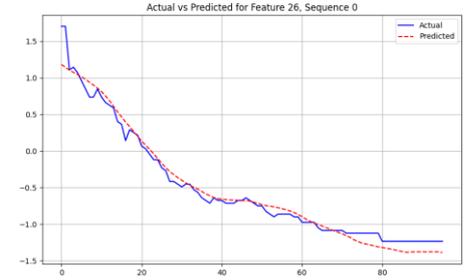

(**a**) T = 96    (**b**) T=96

**Figure 6.** Predicted vs. actual forecasting using our model with L= 96 and T = {96} for PEMS03 and PEMS07 dataset.

## 4. Discussion

One of the recent most effective models for time series forecasting is Dlinear. When we compare our approach to the Dlinear model, we obtain substantial improvements across various datasets as indicated by the results in Table 2. The most significant enhancements are seen in the Weather dataset, with improvements of 18.18% for T=96 and 12.73% for T=192. Notable improvements are also observed in the Illness dataset (22.62% fort T=36) and ETTh2 dataset (11.23% for T=192). These results indicate that our xLSTMTime model consistently outperforms DLinear, especially in complex datasets for varying prediction lengths.

Another notable recent model for time series forecasting is PatchTST. The comparison between our xLSTMTime model and PatchTST reveals a nuanced performance landscape. xLSTMTime demonstrates modest but consistent improvements over PatchTST in several scenarios, particularly in the Weather dataset, with enhancements ranging from 1.03% to 3.36%. The most notable improvements were observed in Weather forecasting at T=96 and T=336, as well as in the ETTh1 dataset for T=720 (1.34% improvement). In the Electricity dataset, xLSTMTime shows slight improvements at longer prediction lengths (T=336 and T=720). However, xLSTMTime also shows some limitations. In the Illness dataset, for shorter prediction lengths, it underperforms PatchTST by 14.78% for T=24, although it outperforms for T=60 by 3.54%. Mixed results were also observed in the ETTh2 dataset, with underperformance for T=336 but better performance at other prediction lengths. Interestingly, for longer prediction horizons (T=720), the performance of xLSTMTime closely matches or slightly outperforms PatchTST across multiple datasets, with differences often less than 1%. This could be attributed to the better long term memory capabilities of the xLSTM approach.

Overall, the comparative analysis suggests that while xLSTMTime is highly competitive with PatchTST, a state-of-the-art model for time series forecasting, its advantages are specific to certain datasets and prediction lengths. Moreover, its consistent outperformance of DLinear across multiple scenarios underscores its robustness. The overall performance profile of xLSTMTime, showing significant improvements in most cases over DLinear and PatchTST, establishes its potential in the field of time series forecasting. Our model demonstrates particular strengths at longer prediction horizons in part due to the long context capabilities of xLSTM coupled with extraction of seasonal and trend information in our implementation.

In comparing the xLSTMTime model with iTransformer, RLinear, PatchTST, Crossformer, DLinear, and SCINet on the PeMS datasets (Table 3), we also achieve superior performance. For instance, in the PEMS03 dataset, for a 12-step prediction, xLSTMTime achieves approximately 9% better MSE, and 5% better MAE. This trend continues across other prediction intervals and datasets, highlighting xLSTMTime's effectiveness in multivariate forecasting. Notably, xLSTMTime often achieves the best or second-best results in almost all cases, underscoring its effectiveness in various forecasting scenarios.

## 5. Conclusions

In this paper, we adapt the recently enhanced recurrent architecture of xLSTM which has demonstrated competitive results in the NLP domain for time series forecasting. Since xLSTM with its improved stabilization, exponential gating and higher memory capacity offer potentially a better deep learning architecture, by properly adapting it to the time series domain via series decomposition, batch and instance normalization, we develop the xLSTMTime architecture for LTSF. Our xLSTMTime model demonstrates excellent performance against state-of-the-art transformer-based models as well as other recently proposed time series models. Through extensive experiments on diverse datasets, the xLSTMTime showed superior accuracy in terms of MSE and MAE, making it a viable alternative to more complex models. We highlight the potential of xLSTM architectures in the time series forecasting arena, paving the way for more efficient and interpretable forecasting solutions, and further exploration using recurrent models.